\documentclass[10pt,twocolumn]{article}

\usepackage[margin=0.72in,columnsep=0.24in]{geometry}
\usepackage[T1]{fontenc}
\usepackage[utf8]{inputenc}
\usepackage{lmodern}
\usepackage{microtype}
\usepackage{amsmath,amssymb}
\usepackage{booktabs}
\usepackage{array}
\usepackage{enumitem}
\usepackage{xcolor}
\usepackage{graphicx}
\usepackage{tikz}
\usetikzlibrary{arrows.meta,positioning,fit}
\usepackage[numbers,sort&compress]{natbib}
\usepackage[hidelinks]{hyperref}
\usepackage{url}
\newcolumntype{L}[1]{>{\raggedright\arraybackslash}p{#1}}

\definecolor{navy}{HTML}{17324D}
\definecolor{blue}{HTML}{2B6CB0}
\definecolor{teal}{HTML}{248B8B}
\definecolor{orange}{HTML}{D97706}
\definecolor{softblue}{HTML}{EAF2FA}
\definecolor{softteal}{HTML}{E7F5F3}
\definecolor{softorange}{HTML}{FFF3DE}
\definecolor{softgray}{HTML}{F3F5F7}

\hypersetup{
  colorlinks=true,
  linkcolor=navy,
  citecolor=blue,
  urlcolor=teal,
  pdftitle={From Score Matrices to Football-Aware Match-State Simulation},
  pdfauthor={Shaopeng Liang}
}
\setlist[itemize]{leftmargin=*,nosep}
\setlist[enumerate]{leftmargin=*,nosep}
\title{\textbf{From Score Matrices to Football-Aware Match-State Simulation:}\\
An Auditable LLM Harness for Exact-Score Reranking}
\author{Shaopeng Liang\\\small Independent Researcher}
\date{5 August 2026}

\begin{document}
\maketitle

\begin{abstract}
Football score forecasting has a useful statistical core and a difficult contextual edge. Dynamic Poisson-family models understand statistical regularities: they estimate current strength, expected goals, and coherent score probabilities. They do not, by themselves, understand football in the semantic sense of roles, tactical matchups, motivation, or the way a first goal changes behaviour. Large language models (LLMs) can reason about those concepts, yet they are neither calibrated probability engines nor guaranteed elite football experts. We therefore treat complementarity, rather than replacement, as the modeling premise. This paper documents four major system iterations. V1 is a dynamic score-driven Dixon--Coles baseline. V2 inserts an LLM residual between two mathematical layers and maps contextual ratings back into $\lambda$. V3 replaces scalar correction with three explicit goal-by-goal simulations over a frozen score-candidate set. V4 adds shared first-breakthrough and post-goal cascade judgments, time-aware stopping, and deterministic tail candidates. The information harness defines every input field, supplies pre-match evidence, and constrains the LLM to an auditable reasoning route.

On an interim chronological replay of the first 150 matches of the 2025--26 English Premier League, V1 score-matrix ordering achieved 10.0\% Top-1 and 26.7\% Top-3 exact-score accuracy. V3 reached 12.0\% and 30.0\%; V4 reached 14.7\% and 30.7\%. V4 increased candidate coverage from 77.3\% to 84.7\%, but no newly added tail candidate became an exact Top-3 hit. The LLM ranking versions did not emit normalized probabilities, so proper scoring rules cannot be computed for them; V1's native 1X2 distribution achieved 0.9878 log loss, 0.5870 Brier score, 0.2095 ranked probability score, and 53.3\% argmax accuracy on the same slice. These results are exploratory rather than a claim of state-of-the-art performance: the slice informed model development, and input isolation cannot rule out outcome knowledge in a closed-weight LLM. The main contribution is therefore an auditable architecture, four clearly separated design iterations, and negative findings that clarify where football simulation does and does not improve score selection.
\end{abstract}

\section{Introduction}

Association-football forecasts have unusually high outcome entropy. A small number of goals, correlated low-score outcomes, and match-state feedback make exact-score prediction especially brittle. Poisson and Dixon--Coles models remain attractive because they produce a complete, internally consistent score distribution rather than only a label \citep{dixoncoles1997}. Dynamic weighting and time depreciation help those models reflect current team strength \citep{ley2018}. Recent distribution-first work similarly argues for predicting the mechanisms that generate scores---for example shot quantity and quality---before simulating outcomes \citep{mendesneves2025}.

The missing layer is often not another generic tabular model. Shortly before kickoff, an analyst wants to interpret whether a reported absence is structurally important, whether a replacement changes ball progression or rest defence, whether a particular winger-fullback matchup is repeatable, and whether the match incentives favour risk or control. These are relational, contextual judgments. An LLM has broad football knowledge and general reasoning ability, but it should not be presumed to possess the consistent specialist judgment of an elite coach, scout, or analyst. The harness is therefore not a passive prompt wrapper: it supplies definitions, reference frames, structured team and player evidence, uncertainty cues, and a required reasoning route. It turns two incomplete components into a new system-level forecaster. A new generation of football benchmarks has begun evaluating LLMs and research agents under pre-match evidence constraints \citep{wang2026}. However, letting an LLM freely predict a score introduces three problems: numerical priors can be ignored, explanations can be post-hoc, and historical simulations can leak results.

This work asks a narrower question:

\begin{quote}
Can an LLM add useful, auditable match-mechanism reasoning while remaining subordinate to a frozen probabilistic score prior?
\end{quote}

The project reached its current form through four major iterations and two interface changes. V1 was purely mathematical. V2 used a ``mathematics + LLM + mathematics'' sandwich: the base model produced $\lambda$, the LLM scored football information, and a learned numerical bridge converted that signal back into $\lambda$. The global residual produced only small, temporally unstable gains. V3 therefore changed the LLM's job from parameter correction to explicit match simulation. V4 retained that interface and made the shared first-goal, stopping, and cascade judgments observable. In V3 and V4, the mathematical model provides the probabilistic prior and candidate geometry; the LLM reads that prior through the harness, constructs causal score paths, and returns a ranking without another formula overwriting its reasoning.

Our contributions are:

\begin{itemize}
  \item an auditable hybrid architecture separating probabilistic state estimation, evidence packaging, LLM reasoning, and deterministic validation;
  \item a documented evolution from feature extraction and scalar residuals to explicit match-state paths, shared root adjudication, and cascade depth;
  \item a chronological replay protocol that freezes prompts and responses before outcome scoring, with per-match input/output archives and hashes; and
  \item an honest interim benchmark including proper-score baselines, paired comparisons, failure slices, and limitations of historical LLM evaluation.
\end{itemize}

\section{Related Work}

Classical score models estimate team attack and defence effects and construct a joint distribution for home and away goals. Dixon and Coles introduced a dynamic Poisson-regression framework and a low-score correction for the cells most affected by independence assumptions \citep{dixoncoles1997}. Ley et al. compared strength-based models under match-importance and time-depreciation weights and found independent and bivariate Poisson formulations competitive under ranked probability score (RPS) evaluation \citep{ley2018}. More recent comparisons report that changing the machine-learning family or feature set often yields only modest improvements over careful Poisson baselines \citep{fischer2024}.

Distribution forecasting is important because a single exact score hides uncertainty. Mendes-Neves et al. combine Elo information, learned shot distributions, and Monte Carlo simulation to obtain match distributions \citep{mendesneves2025}. Proper scoring rules reward an honest predictive distribution rather than a fortunate hard label \citep{gneiting2007}. We report log loss, Brier score, and RPS for V1, while noting that the relative suitability of RPS and local ignorance/log scores for football remains debated \citep{wheatcroft2019}.

WorldCupArena evaluates LLMs and deep-research agents using evidence available before kickoff and reports result, exact-score, and graded scoreline metrics \citep{wang2026}. Our focus is complementary: instead of benchmarking unconstrained model knowledge, we study how an LLM can be inserted after a score-distribution model under a fixed interface, and we preserve its complete inputs, outputs, and validation decisions.

\section{System Architecture}

\begin{figure*}[t]
\centering
\resizebox{0.95\textwidth}{!}{%
\begin{tikzpicture}[
  node distance=8mm and 7mm,
  box/.style={rounded corners=2pt, draw=navy, line width=0.7pt, align=center, minimum height=12mm, text width=29mm, font=\small},
  arr/.style={-{Latex[length=2mm]}, line width=0.7pt, draw=navy},
  note/.style={font=\scriptsize, align=center, text width=29mm}
]
\node[box,fill=softblue] (hist) {Past match results\\chronological state};
\node[box,fill=softblue,right=of hist] (prior) {V1 statistical prior\\$\lambda_H,\lambda_A$};
\node[box,fill=softorange,right=of prior] (pool) {Score matrix\\frozen candidates};
\node[box,fill=softteal,right=of pool] (harness) {Annotated pre-match\\evidence harness};
\node[box,fill=softteal,right=of harness] (sim) {V4 match simulation\\root, cascade, paths};
\node[box,fill=softorange,below=10mm of sim] (rank) {Validated ranking\\Top-1 / Top-3};
\node[box,fill=softgray,left=of rank] (freeze) {Freeze prediction\\prompt, output, hashes};
\node[box,fill=softgray,left=of freeze] (score) {Reveal outcome\\and evaluate};

\draw[arr] (hist) -- (prior);
\draw[arr] (prior) -- (pool);
\draw[arr] (pool) -- (harness);
\draw[arr] (harness) -- (sim);
\draw[arr] (sim) -- (rank);
\draw[arr] (rank) -- (freeze);
\draw[arr] (freeze) -- (score);

\node[note,below=2mm of prior] {Probability owner};
\node[note,above=1mm of harness] {No result fields};
\node[note,below=2mm of rank] {No post-hoc formula};
\end{tikzpicture}
}
\caption{The V4 pipeline. The forward path is strictly one-way: predictions are frozen before a separate evaluator reads outcomes. Temporal input isolation does not exclude outcome memory inside a closed LLM.}
\label{fig:architecture}
\end{figure*}
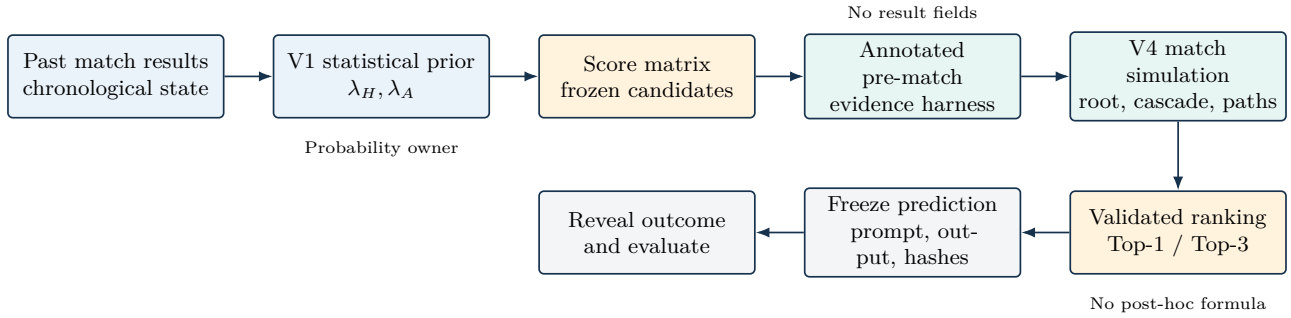

Figure~\ref{fig:architecture} shows the separation of responsibilities. V1 owns numerical probability. The harness owns evidence provenance and semantics. V4 owns structured match simulation. A deterministic validator owns schema compliance and ensures that the returned scores belong to the allowed candidate pool.

\subsection{V1: dynamic probabilistic prior}

V1 is a score-driven (generalized autoregressive score, GAS) attack--defence model. For team $i$, the pre-match attack and vulnerability states decay toward zero as
\begin{align}
 a^-_{i,t}&=a_{i,t^-}\exp\!\left(-\log 2\,\frac{\Delta t}{h}\right),\\
 v^-_{i,t}&=v_{i,t^-}\exp\!\left(-\log 2\,\frac{\Delta t}{h}\right).
\end{align}
where $h=1440$ days in the selected V1 configuration. For home team $H$ and away team $A$,
\begin{equation}
 \lambda_H=\mu_H\exp(a^-_H+v^-_A),\qquad
 \lambda_A=\mu_A\exp(a^-_A+v^-_H).
\end{equation}
The states are updated after each date-batch using clipped, standardized score innovations. The selected gains are $g_a=0.04$ for attack and $g_v=0.02$ for vulnerability. This combines gradual forgetting with rapid online reaction to surprising goals; the half-life alone does not determine responsiveness.

Independent Poisson masses are multiplied by the Dixon--Coles low-score correction $\tau_{xy}(\lambda_H,\lambda_A,\rho)$, with $\rho=-0.05$, and renormalized:
\begin{equation}
 p(x,y)=\frac{\operatorname{Pois}(x;\lambda_H)\operatorname{Pois}(y;\lambda_A)\tau_{xy}}{Z}.
\end{equation}
Aggregating matrix cells gives native home/draw/away probabilities. Sorting cells gives exact-score candidates. These are different decision problems: the modal score's match result need not equal the largest aggregate 1X2 class.

V1 was selected using 18,665 matches from 2015--16 through 2024--25, covering five top domestic leagues and the UEFA Champions League, Europa League, and Conference League. Seasons through 2021--22 formed the training window; 2022--23 through 2024--25 formed validation. The selected configuration achieved 0.9855 1X2 log loss, 0.5876 Brier score, 0.2004 RPS, and 52.8\% accuracy on 5,330 domestic validation matches. The 2025--26 season was not used in V1 parameter selection, but it is used in the retrospective LLM development described below.

\subsection{The pre-match evidence harness}

Each match packet explains both values and their meanings. It contains:

\begin{itemize}
  \item kickoff and evidence cutoff; home/away identity and competition context;
  \item V1 $\lambda_H$, $\lambda_A$, full matrix semantics, and ranked candidates;
  \item table position, season objective, recent form, goals, formations, rest, and schedule congestion;
  \item player starts, minutes, role, position, expected lineup, rotation, replacement, and availability information;
  \item pre-match reporting and one or more key tactical matchups; and
  \item explicit reminders that reporting is uncertain evidence rather than ground truth.
\end{itemize}

The LLM must use football reasoning, including source authority and contextual plausibility, rather than literal keyword matching. The harness does not ask the LLM to prove whether a report was ultimately correct. Its task is to judge the information available at prediction time and explain how much it should matter. Appendix~\ref{app:prompts} documents the prompt contracts and representative success and failure cases.

\section{Why the Model Changed}

Table~\ref{tab:evolution} summarizes four major research iterations. Internal engineering builds are intentionally collapsed into these public model versions; the paper studies changes of representation rather than a sequence of prompt patches.

\begin{enumerate}
  \item \textbf{V1: mathematics alone.} A dynamic score model learns team strength and produces a coherent probability matrix, but football meaning enters only indirectly through past goals.
  \item \textbf{V2: mathematics + LLM + mathematics.} The LLM interprets current football information; one residual adapter compresses that judgment and pushes it back into $\lambda$.
  \item \textbf{V3: mathematics + LLM path simulation.} The statistical model supplies the prior and candidate geometry; the LLM explicitly simulates three goal-by-goal paths and reranks candidates without changing $\lambda$.
  \item \textbf{V4: shared-root and cascade simulation.} The path model adds a common first-breakthrough decision, time-aware stopping, post-goal response and expansion, and controlled tail candidates.
\end{enumerate}

The later iterations do not claim that the LLM is a top football expert. They make broad football knowledge operational by supplying missing specialist context and forcing the reasoning through auditable states. In this sense, the harness is part of the predictive model itself.

\begin{table*}[t]
\centering
\small
\begin{tabular}{L{0.06\textwidth}L{0.19\textwidth}L{0.29\textwidth}L{0.29\textwidth}}
\toprule
Version & LLM's job & Hypothesis & What the experiment taught us \\
\midrule
V1 & None & Dynamic team states plus a corrected score matrix form a stable probabilistic backbone. & Retained as the numerical baseline and probability owner. \\
V2 & Produce four relative attack/defence ratings; one learned $\kappa$ maps them back into $\lambda$. & A compact numerical adapter can connect contextual football judgment to the probability model. & Pooled gains were tiny and temporally unstable. One residual cannot express evidence whose value is asymmetric or state-dependent. \\
V3 & Build three explicit goal-by-goal paths from 0--0 and rerank a frozen Top-10. & Causal match simulation can discipline score selection without editing $\lambda$. & Interpretability improved, but candidate coverage became a hard ceiling and stopping assumptions were exposed as failure modes. \\
V4 & Judge first breakthrough, remaining-time stopping, and post-goal cascade over an expanded candidate pool. & Repeatable scoring channels and the trailing side's response capacity should control score depth. & Top-3 improved descriptively, but 0--0 remained unsolved and added tail candidates contributed no new exact hit. \\
\bottomrule
\end{tabular}
\caption{Four public research versions and the design reasoning behind each transition. Internal mappings are V1/V3.2, V2/V5.3, V3/V7.25, and V4/V7.28.}
\label{tab:evolution}
\end{table*}

\subsection{V2: a clean scalar residual was too uniform}

V2 assigns four LLM values---home attack, home defence, away attack, and away defence---relative to each team's own mathematical baseline centred at 10. A single learned $\kappa$ maps rating differences into capped log-$\lambda$ adjustments. This architecture preserves a normalized probability distribution and is easy to evaluate, but it compresses a conditional football story into one scalar bridge.

The final V2 review combined three chronological segments totalling 320 matches. Pooled 1X2 log loss improved by 0.1365\%, but the last 99-match segment regressed by 0.2387\% and lost three hard 1X2 picks. The result suggested that a global $\kappa$ was trying to do incompatible jobs. Ordinary evidence should produce almost no movement; decisive absences or tactical mismatches may deserve more; and a first goal can change the sign or magnitude of later effects. These are conditional transitions, not one residual.

\subsection{V3: from parameter correction to explicit paths}

V3 changes the interface rather than adding scalar features. The LLM receives V1's $\lambda$ values, score-matrix semantics, frozen Top-10, and annotated pre-match evidence. It must construct three distinct paths from 0--0. Each transition is one home goal, one away goal, or a stop. After every goal, it reassesses the leader's risk reduction, the trailer's pressure and actual response capacity, newly exposed space, fatigue, substitutions, and key matchups. The three terminal scores become the returned Top-3. The LLM cannot edit $\lambda$, invent an arbitrary score, or apply a post-output weighting formula.

\subsection{V4: shared root, stopping, and cascade}

V4 retains V3's explicit paths and adds the state variables needed to control their depth. Before the paths diverge, one shared root judges whether the 0--0 equilibrium is likely to break. Each node records score, estimated time band, remaining time, and stop-versus-continue evidence. After the first goal, one shared cascade assessment asks whether the scoring mechanism is repeatable, whether the trailing side can actually respond, and whether the leading side can exploit newly opened space. The cascade may close the game, leave openness roughly unchanged, or open it bilaterally or asymmetrically.

The deterministic builder also appends up to five matrix-derived tail anchors: a high-scoring draw, high-scoring home and away wins, and home and away runaway scores. These are optional candidates rather than mandatory selections. V4 still cannot alter $\lambda$ or invent scores outside the frozen pool.

\section{Experimental Protocol}

\subsection{Interim development benchmark}

The reported V4 benchmark is the first 150 chronological fixtures of the 2025--26 English Premier League, grouped into 15 league rounds of 10 matches. The slice is deliberately presented as an \emph{interim development benchmark}. The first 100 matches informed V4's design, so the 150-match result is not an untouched final test. Although later fixtures were also replayed during development, they are outside the main benchmark requested for this draft and cannot be retroactively recovered as unseen data. A credible generalization claim requires a new future season or competition frozen before outcomes.

For each round, up to 10 matches were processed concurrently but independently. Each match used a fresh LLM context. Prompt, source packet, structured response, schema version, model identifier, and content hashes were written and synchronized before a separate evaluator opened the result file. The same date-batch was then supplied to V1's online update before the following round. This prevents target-result fields from entering the prompt and prevents within-round updates.

This protocol is \emph{temporal input isolation}, not a proof of full non-contamination. The closed-weight reasoning model was executed after the historical season; its unknown pretraining or retrieval history could contain match outcomes. The experiment did not probe parametric memorization. Consequently, the benchmark validates the software and reasoning interface more strongly than it validates an unbiased LLM forecasting edge.

\subsection{Metrics}

We report exact-score Top-1 and Top-3 accuracy. We also map the Top-1 score to home/draw/away, denoted \emph{score-derived 1X2 accuracy}. For V1 we separately report \emph{native 1X2 argmax accuracy}, obtained by summing the whole score matrix by outcome before taking the maximum.

For a three-class probability vector $p=(p_H,p_D,p_A)$ and observed one-hot vector $o$, we use
\begin{align}
 \text{LogLoss} &= -\log p_y,\\
 \text{Brier} &= \sum_{k\in\{H,D,A\}}(p_k-o_k)^2,\\
 \text{RPS} &= \frac{1}{2}\sum_{j=1}^{2}\left(\sum_{k=1}^{j}p_k-\sum_{k=1}^{j}o_k\right)^2.
\end{align}
Lower is better. V3 and V4 return only an ordering, not a normalized posterior, so assigning them arbitrary rank weights would not produce valid proper scores.

\section{Results}

\subsection{First-150 comparison}

Table~\ref{tab:mainresults} gives the central benchmark. V4 improves Top-1 exact score by 4.7 percentage points over V1's score-cell mode and Top-3 by 4.0 points. Its score-derived 1X2 accuracy is 18 points higher. Paired exact McNemar tests are not significant for Top-1 ($p=0.2295$) or Top-3 ($p=0.3075$); the score-derived 1X2 difference is significant ($p=0.0024$). This last comparison must not be confused with V1's native 1X2 decision, which reaches 53.3\% and remains above V4's score-derived 50.0\%.

\begin{table}[t]
\centering
\scriptsize
\begin{tabular}{@{}lrrr@{}}
\toprule
Model & Top-1 & Top-3 & Score-1X2 \\
\midrule
V1 score order & 15 (10.0) & 40 (26.7) & 48 (32.0) \\
V3 path simulation & 18 (12.0) & 45 (30.0) & 71 (47.3) \\
V4 cascade simulation & \textbf{22 (14.7)} & \textbf{46 (30.7)} & \textbf{75 (50.0)} \\
\bottomrule
\end{tabular}
\caption{First 150 EPL matches. Counts are followed by percentages in parentheses. ``Score-1X2'' is the result implied by the Top-1 exact score.}
\label{tab:mainresults}
\end{table}

\begin{table}[t]
\centering
\small
\begin{tabular}{lr}
\toprule
V1 native probability metric & Value \\
\midrule
1X2 argmax accuracy & 80/150 (53.33\%) \\
Log loss & 0.987803 \\
Brier score & 0.586970 \\
Normalized RPS & 0.209451 \\
Mean included matrix mass & 0.996674 \\
\bottomrule
\end{tabular}
\caption{Proper-probability evaluation of V1 on the same first-150 slice. The truncated score matrix was renormalized per match.}
\label{tab:proper}
\end{table}

The contrast between 32.0\% score-derived and 53.3\% native 1X2 accuracy is not a bug in matrix aggregation. The most likely individual score cell is often 1--1 or a narrow result, while the sum of many home-win cells can exceed the diagonal. A single modal score is therefore a poor substitute for a categorical probability decision.

\subsection{First-100 V3-to-V4 comparison}

The first-100 slice provides a common descriptive comparison, although it is also the slice that motivated V4. Relative to V3, V4 added one Top-1 hit, three Top-3 hits, and two score-derived 1X2 hits. These counts are descriptive development results rather than independent evidence.

\begin{table}[t]
\centering
\small
\begin{tabular}{lrrr}
\toprule
Model & Top-1 & Top-3 & Score-1X2 \\
\midrule
V3 & 13 & 29 & 47 \\
V4 & \textbf{14} & \textbf{32} & \textbf{49} \\
\bottomrule
\end{tabular}
\caption{Hits out of 100 matches on the shared development slice.}
\label{tab:ablation}
\end{table}

\subsection{Candidate ceiling and tail failure}

The true score appeared in V1's Top-10 for 116 of 150 matches (77.3\%). V4's deterministic anchors expanded theoretical coverage to 127 matches (84.7\%), an increase of 11. Yet an added anchor entered the final Top-3 in only 3 matches, and none was an exact hit. Thus V4 improved the \emph{availability} of tail scores but did not learn how to promote them.

There were 43 matches with at least four total goals and 24 with a margin of at least three goals. V4's Top-3 exactly matched 4 and 3 of these, respectively. By contrast, among 14 actual 1--1 results, V4 placed 1--1 first four times and inside Top-3 ten times. None of the eight 0--0 results appeared in Top-3. The model is much better at choosing among central low-score outcomes than recognizing a genuine no-breakthrough state or a far tail.

\subsection{Reasoning-node diagnostics}

V4 required an explicit root verdict, but all 150 cases selected \texttt{first\_goal\_more\_likely}. This is a useful negative result: exposing an implicit decision creates an auditable variable, but does not create calibration. In the first 100 matches, V4 labelled 72 cascades as \texttt{opens\_game}, 21 as neutral, and only 7 as closing the game. The open-game group had lower Top-3 accuracy than the two other groups combined, suggesting a generic simulation bias toward escalation.

The large-margin adjustment was more promising. In the first 100 matches, four fixtures received a value of at least 2.0; three truly ended with a margin of at least three goals. This is high precision but low recall (3 of 15 tail events) and was not yet mapped into a trained ranking rule. It should be treated as a candidate signal, not evidence of a solved tail problem.

\section{Discussion}

\subsection{What the LLM appears to add}

The strongest empirical case is not that the LLM replaces the probability model. V1's native 1X2 probability decision remains stronger than V4's score-derived result. Rather, the combined system changes which \emph{specific score cells} are made visible to a human and supplies a structured causal simulation for that ordering. Relative to V1's score-cell mode, Top-1 and Top-3 exact-score rates improve descriptively. The explicit tree also makes errors inspectable: one can distinguish an incorrect first-breakthrough judgment, an incorrect reaction to a lead, a correct mechanism with insufficient candidate depth, and a candidate-generation failure.

The result supports a division of epistemic labour. V1 is strong where the task is aggregation: many historical observations, time decay, coherent probabilities, and repeatable arithmetic. It is weak where the task is semantic: why a nominal replacement changes build-up, why a matchup creates a repeatable channel, or why league position changes risk appetite. The LLM has the opposite profile. It can connect those concepts and simulate conditional consequences, but its knowledge is broad rather than consistently expert, and its raw numerical confidence is not calibrated. The harness is the engineered bridge that constrains each component to the work it performs best.

The evolution also supports a broader modeling principle. Context value is heterogeneous. A routine availability report should not receive the same global residual weight as the loss of a unique ball-progressor, nor should a first-half absence have the same effect after a game state changes. V2 treated context too much like a fixed tabular feature. V3 and V4 represent it as conditional evidence attached to transitions.

\subsection{What remains unresolved}

First, V3 and V4 produce rankings, not probabilities. They cannot yet improve or even be compared on proper scoring rules. A future version should output a normalized posterior over the candidate set and reserved tail mass, then be calibrated on a development window and frozen before a true forward test.

Second, V4's root and cascade fields are semantically rich but poorly calibrated. The all-positive first-goal verdict cannot identify 0--0; the frequent open-game label does not reliably select high totals. These signals need an empirical mapping, but returning to one global $\kappa$ would repeat the V2 error. A hierarchical calibration layer should condition on signal type, confidence, information value, game state, and remaining time, with shrinkage toward no adjustment.

Third, candidate expansion alone is insufficient. Tail candidates existed but were rarely ranked. The next experiment should separate two questions: whether the candidate generator covers the outcome and whether the reasoner promotes it. Coverage and ranking should be scored independently.

Fourth, exact score is intrinsically noisy. Future reports should add a graded scoreline distance metric, total-goal and goal-difference distributions, calibration plots, and decision-quality measures alongside exact hits. WorldCupArena's fine-grained scoreline evaluation offers one relevant direction \citep{wang2026}.

\section{Validity, Reproducibility, and Responsible Use}

\paragraph{Development reuse.} The first 100 matches influenced V4. The first-150 benchmark is therefore an engineering benchmark, not confirmatory evidence. Statistical $p$-values are descriptive and do not repair adaptive model selection.

\paragraph{Model-memory contamination.} The LLM was run after the target season. Inputs exclude target scores and are frozen before scoring, but a closed model may have memorized outcomes. The current audit cannot rule this out. A future live, preregistered evaluation is mandatory.

\paragraph{Stochasticity.} LLM ranking can vary across runs. The present benchmark reports one archived run per version. Replicated inference with fixed sampling metadata and paired uncertainty intervals is needed.

\paragraph{Evidence quality.} Pre-match reports can be incomplete, revised, strategically misleading, or wrong. The method intentionally asks the LLM to reason under uncertainty, but source reliability remains a model input rather than verified truth.

\paragraph{Reproducibility.} Code, version contracts, prompts, structured outputs, and review logs are anchored at Git tag \nolinkurl{football-forecast-v7.28-frozen-2026-08-04}. The internal LLM run identifier is \texttt{gpt-5.6-sol} with high reasoning effort. Because this is a closed-weight service model, exact external reproduction is not guaranteed even with identical prompts.

\paragraph{Responsible use.} This research studies football forecasting and human-readable AI reasoning. It does not provide wagering recommendations. Exact-score outputs are uncertain hypotheses, and reported historical accuracy should not be interpreted as expected financial return.

\section{Conclusion}

This project began with a purely statistical model and then moved to a conventional hybrid idea: estimate two scoring rates, let an LLM interpret context, and convert that interpretation back into a small mathematical correction. The experiments showed why that ``mathematics + LLM + mathematics'' sandwich was incomplete. LLM player-state judgments could be useful while their aggregate $\lambda$ effect was wrong; a global $\kappa$ could show a tiny pooled gain while failing temporally; and a richer prompt could make reasoning easier to inspect without making the root decision more discriminative.

V4 therefore represents a different model class from V2's residual sandwich: ``mathematics + LLM''. The statistical component keeps ownership of probability and candidate generation. The LLM contributes football semantics and conditional simulation. The harness supplies the domain scaffolding neither component has alone and makes the joined reasoning executable and auditable. On the first-150 development replay, V4 reaches 14.7\% Top-1 and 30.7\% Top-3 exact-score accuracy, but does not outperform V1's native 1X2 probability decision, solve 0--0, or convert new tail coverage into tail hits.

The main result is a methodology for disciplined hybridization, not a headline accuracy claim. The next scientifically meaningful step is a preregistered, live forward test in which the LLM outputs a calibrated posterior and neither developers nor model services have access to outcomes at prediction time.

\section*{Acknowledgments}

The author used an LLM-based coding and reasoning assistant to help implement experiments, audit artifacts, and draft this manuscript. The human author is responsible for all claims, numerical results, citations, and the final submitted text.

\bibliographystyle{plainnat}
\bibliography{references}

\appendix
\section{Prompt Design and Representative Cases}
\label{app:prompts}

This appendix explains what the LLM was actually asked to do at each stage. The production prompts were written in Chinese and are archived with the code. To keep the paper readable, the boxes below are faithful English contract summaries rather than full verbatim copies. Each case reproduces values from an archived input/output artifact; the result is mentioned only in the post-hoc case review. Several cases come from development splits and therefore illustrate behaviour rather than independent performance.

\subsection{Prompt design across versions}

The prompt evolved along four axes: the object being predicted, the numerical authority granted to the LLM, the explicitness of the reasoning route, and the output that deterministic code could validate. Table~\ref{tab:prompt-evolution} makes those changes explicit.

\begin{table*}[t]
\centering
\footnotesize
\begin{tabular}{L{0.06\textwidth}L{0.15\textwidth}L{0.23\textwidth}L{0.20\textwidth}L{0.16\textwidth}}
\toprule
Version & Prediction object & Required reasoning & Structured output & Hard constraint \\
\midrule
V1 & Score distribution; no LLM & Chronological team-state update $\rightarrow$ $\lambda_H,\lambda_A$ $\rightarrow$ corrected score matrix & Normalized score and 1X2 probabilities & Target match cannot update its own prior \\
V2 & Match-day residual relative to V1 & Baseline $\rightarrow$ lineup and roles $\rightarrow$ matchup, schedule, motivation $\rightarrow$ four relative ratings & Four attack/defence ratings, confidence, causal rationale & Ratings are centred at each team's own 10.0 baseline; deterministic $\kappa$ alone changes $\lambda$ \\
V3 & Three causal terminal scores and full candidate ordering & Three paths from 0--0; after each goal reassess risk, response capacity, space, fatigue, substitutes, and matchup effects & Three score paths, mechanisms, and a Top-10 permutation & Terminals must equal Top-3; no arbitrary score, $\lambda$ edit, or post-output formula \\
V4 & Shared root and cascade plus timed paths over an expanded pool & First breakthrough $\rightarrow$ score/time stopping $\rightarrow$ repeatability, trailing response, leading expansion, and cascade depth & Root verdict, cascade class, tail-risk signals, timed paths, and full-pool permutation & All paths share root and cascade; tail anchors are optional and require match-specific support \\
\bottomrule
\end{tabular}
\caption{Executable contracts for the four public versions. The principal change is from estimating context as a static residual to simulating conditional match states.}
\label{tab:prompt-evolution}
\end{table*}

The public versions map to the frozen internal implementations V3.2, V5.3, V7.25, and V7.28. Raw prompt sources, response schemas, validators, and complete match-level inputs and outputs are archived independently; the public naming keeps the scientific argument readable without changing those immutable artifacts.

\subsection{V2: relative ratings and a numerical bridge}

\paragraph{Prompt contract.} V2 begins with a crucial division of authority: ``The mathematical baseline owns absolute worldwide team strength.'' The LLM must judge only how today's attack and defence differ from the rolling state already represented by $\lambda_H$ and $\lambda_A$. A value of 10.0 means 100\% of that team's own baseline, not a universal club rating. The required route is: interpret the baseline; infer lineup and roles; compare replacement functions; integrate the opponent-specific matchup, schedule, fitness, motivation, and expected game state; then derive residual indices with an explicit causal bridge. A causal-ownership ledger prevents one mechanism from being counted from both teams' perspectives. Deterministic code alone maps final indices into capped log-$\lambda$ changes through $\kappa$.

\paragraph{Case V2-C1: Newcastle United versus Bournemouth.} The V1 prior was $(\lambda_H,\lambda_A)=(1.649,1.403)$, with a modal 1--1 and away-win probability 0.322. The LLM described Newcastle as needing to press at home while Bournemouth was structurally better suited to transition into the space behind the full-backs. Gordon remained Newcastle's principal outlet, but Bournemouth had more simultaneous progression and finishing contributions. It returned home attack/defence indices $(9.4,9.4)$ and away indices $(10.5,10.2)$ with 0.69 confidence. Under the fitted $\kappa=2.0$ adapter, the rates became $(1.400,1.714)$ and away-win probability rose to 0.445. The actual score was 1--2, so the adjustment moved the result distribution in the correct direction; nevertheless, the modal exact score remained 1--1. The case captures both the attraction and the limitation of the sandwich architecture: the LLM found a plausible matchup signal, while a single residual bridge could not fully express the score path.

\paragraph{Why the prompt changed next.} Across larger samples, the scalar bridge produced small, unstable gains. The system therefore stopped asking the LLM to be a noisy numerical parameter estimator and asked it instead to rank concrete score scenarios under explicit constraints.

\subsection{V3: explicit score paths}

\paragraph{Prompt contract.} V3 receives the frozen V1 $\lambda$ values, score matrix, Top-10 candidates, and complete pre-match packet. The prompt explains that the matrix is an unconditional prior and plausibility map, not a final answer. The LLM must expose three paths from 0--0. Each step increments only the home or away score by one, and every goal triggers a fresh judgment of tactical risk, space, fatigue, substitutes, response capacity, and matchup effects. The three terminals must be distinct and must equal the returned Top-3.

\paragraph{Case V3-C1: Brentford versus Manchester City.} The input rates were $(1.315,1.884)$ and V1's first three cells were 1--1, 1--2, and 0--2. V3 returned 1--2, 0--2, and 1--3. Its three stories were: City score, Brentford equalize, and City find the winner; City score twice while suppressing the counter; or Brentford equalize before its higher line leaves space for a third City goal. The actual score was 0--1. V3 corrected the broad direction but missed the exact score because every leading path assumed either a Brentford response or a second City goal. The explicit simulation made the missing 0--1 stopping branch visible.

This case is taken from the dedicated V3 audit artifact, not substituted into the first-150 score table. Its purpose is to show how an exact-score error becomes a specific path error.

\subsection{V4: shared root, stopping, cascade, and tail candidates}

\paragraph{Prompt contract.} V4 begins with one shared root before the three paths. The LLM must separately state evidence for no breakthrough and for at least one goal, distinguish theoretical channels from recently verified ones, and report a root verdict, information value, and confidence. Each path node then records current score, estimated time band, and remaining time; it must compare \emph{score holds} with \emph{another goal} before choosing a scorer. After the first goal, a shared cascade judgment identifies the mechanism and asks whether it is repeatable. The LLM separately judges the trailing side's \emph{actual} response capacity and the leader's ability to reuse the mechanism or attack newly opened space. It distinguishes bilateral opening from one-sided expansion. The output includes \path{closes_game}, \path{neutral}, or \path{opens_game}; high-total and large-margin signals; runaway direction; timed paths; information value; and confidence.

The deterministic input builder also appends at most five matrix-derived anchors: the highest-probability high-scoring draw, high-scoring home and away wins, and home and away runaway scores. The prompt repeatedly warns that these are candidates rather than mandatory tail selections.

\paragraph{Case V4-C1: Arsenal versus Nottingham Forest.} The root favoured a first goal, noting Arsenal's recent striker and delivery output, Forest's five goals conceded in three matches, and the conditional risk of a higher defensive line. The cascade classified the first goal as \path{opens_game}, with high-total adjustment $+1.3$, large-margin adjustment $+2.1$, home runaway direction, 0.80 information value, and 0.73 confidence. Its mechanism was asymmetric: Forest had a real Wood/Gibbs-White response route, but chasing the game was more likely to expose the same space Arsenal had already used. The leading-side expansion label was \path{strong}, while the trailing response was \path{opens_more_space}. The Top-3 became 2--0, 3--0, 2--1; the actual score was 3--0, a Top-2 hit.

The result should be interpreted carefully. The correct 3--0 cell already belonged to V1's Top-10, so the hit demonstrates successful cascade-driven \emph{reranking}, not a direct success of the newly appended tail anchors. This distinction is exactly why V4 audits candidate coverage and ranking separately.

\subsection{Cross-case lessons}

The cases reveal a consistent pattern. Prompt structure can make a useful football judgment more explicit, but cannot guarantee that the judgment is calibrated or transferred correctly:

\begin{itemize}
  \item V2 identified a useful matchup direction, but the global numerical bridge compressed a conditional story into one residual.
  \item V3 made score selection auditable by exposing the missing 0--1 stopping path in a concrete failed case.
  \item V4 could promote a plausible large-margin score when the cascade mechanism was specific, yet its shared root remained biased toward a first goal and its added tail anchors produced no exact Top-3 hit over the first 150 matches.
\end{itemize}

For future versions, each new prompt field should therefore satisfy two tests: it must make a previously hidden football decision observable, and that observable decision must show calibration or ranking value on a genuinely forward sample. Interpretability is necessary for diagnosis, but it is not itself predictive improvement.

\section{Audit Checklist}

For each target match, a valid archived run must satisfy all of the following:

\begin{enumerate}
  \item the packet cutoff precedes kickoff and contains no target result field;
  \item V1 prediction is generated before the match update;
  \item prompt and response are stored in full with hashes;
  \item the response passes the version-specific JSON schema;
  \item every returned score belongs to the frozen candidate pool;
  \item three Top-3 terminals are distinct and connected to explicit paths;
  \item prediction artifacts are synchronized before the result reader executes; and
  \item all matches on one date are predicted before any same-date state update.
\end{enumerate}

\section{Version Contracts}

V1 owns normalized probability. V2 alone may modify $\lambda$, and only through its deterministic $\kappa$ adapter. V3's path node permits three transitions: next home goal, next away goal, or stop. V4 places one shared first-breakthrough object before the paths, records score and time at every node, places one shared cascade object after the root, and expands the candidate pool deterministically. Earlier implementations and outputs remain immutable; no later field is silently backported.

\section{Planned Confirmatory Evaluation}
\enlargethispage{4\baselineskip}

A future protocol should be registered before fixtures begin. It should freeze: leagues and rounds; evidence cutoff; LLM model/version and sampling settings; prompt and schema hashes; candidate-generation rules; probability-calibration procedure; primary metric; and stopping rule. The primary outcome should be a proper score on normalized probabilities. Exact Top-1/Top-3, graded scoreline distance, 0--0 recall, high-total recall, and candidate coverage should be secondary diagnostics. All predictions should be timestamped and publicly committed before kickoff.

\end{document}